\documentclass[a4paper,twoside,french]{article}
\usepackage{orasis}
\usepackage[utf8]{inputenc}
\usepackage[T1]{fontenc}
\usepackage[english]{babel}
\usepackage{times}

\usepackage{graphicx}
\usepackage{xcolor}
\usepackage{amsmath}  

\usepackage{booktabs}

\begin{document}
%%%%%Pas de date
\date{}
%%%%% Titre gras 14 points
\title{\Large\bf Fast 3D point clouds retrieval for Large-scale 3D Place Recognition}
%%%%% Si auteur unique
%\author{L. Auteur \\
%%  Son institut \\
%%  Son addresse \\
%%  Son email}
%%%% pour deux auteurs
%%%%\author{\begin{tabular}[t]{c@{\extracolsep{8em}}c}
%%%% pour trois auteurs
\author{\begin{tabular}[t]{c@{\extracolsep{6em}}c@{\extracolsep{6em}}c}
%%%% pour quatre auteurs
%%%%\author{\begin{tabular}[t]{c@{\extracolsep{4em}}c@{\extracolsep{4em}}c@{\extracolsep{4em}}c}
%%%%pour plus débrouillez-vous !
  Chahine-Nicolas Zede & Laurent Caraffa &  Valérie Gouet-Brunet \textbf{}   \\
\end{tabular}
{} \\
 \\
LASTIG, IGN-ENSG, Gustave Eiffel University, 77420 Champs-sur-Marne, France   \\
{} \\
 \\
%Mon adresse complète \\
chahine-nicolas.zede@ign.fr, laurent.caraffa@ign.fr, valerie.gouet@ign.fr\
}

\maketitle
%%%%  Pas de numérotation sur la page de titre
\thispagestyle{empty}
\subsection*{Résumé}
Rechercher un nuage de points 3D est une tâche difficile qui consiste à identifier, au sein d'une base de référence, les nuages les plus similaires à un nuage requête donné. Les méthodes actuelles se concentrent principalement sur la description de ces nuages de points comme indicateur de similarité. Dans ce travail, nous nous focalisons sur l'accélération de la recherche en adaptant le \textit{Differentiable Search Index} (DSI), basé sur les transformers et initialement conçu pour la recherche d'information textuelle, à la recherche dans les nuages de points. Notre approche génère des identifiants 1D, associés à des descripteurs 3D, permettant ainsi une recherche très efficace en temps constant. Pour adapter cette approche aux nuages de points, nous utilisons les \textit{Vision Transformers} qui associent les descripteurs multidimensionnels à leurs identifiants 1D, tout en intégrant un encodage positionnel et sémantique. Les performances de notre proposition sont évaluées face à l'état de l'art sur un jeu de données public, dans le contexte de la  reconnaissance de lieux, en termes de qualité et rapidité des réponses retournées.

\subsection*{Mots Clef}
Recherche 3D, recherche à grande échelle, descripteurs 3D, index de recherche différentiable, Reconnaissance de lieux basée sur le LiDAR.

\subsection*{Abstract}
{\em
Retrieval in 3D point clouds is a challenging task that consists in retrieving the most similar point clouds to a given query within a reference of 3D points. Current methods focus on comparing descriptors of point clouds in order to identify similar ones. Due to the complexity of this latter step, here we focus on the acceleration of the retrieval by adapting the Differentiable Search Index (DSI), a transformer-based approach initially designed for text information retrieval, for 3D point clouds retrieval. Our approach generates 1D identifiers based on the point descriptors, enabling direct retrieval in constant time. To adapt DSI to 3D data, we integrate Vision Transformers to map descriptors to these identifiers while incorporating positional and semantic encoding. The approach is evaluated for place recognition on a public benchmark comparing its retrieval capabilities against state-of-the-art methods, in terms of quality and speed of returned point clouds.
}
\subsection*{Keywords}
3D retrieval, Scalable retrieval, 3D point cloud descriptors, Differentiable Search Index (DSI), LiDAR-based place recognition

\section{Introduction}
%approfondir la présentation de l'introduction (détailler plus les applications potentielles ?) et surtout de l'état de l'art en détaillant plus les approches (section 2.1).

Without any initial information on geolocation, the process of geolocalization from visual content typically begins with content-based indexing and retrieval \cite{dubey2021decade}. In this work, we focus on 3D point clouds data, potentially obtained from LiDAR sensors, Structure-from-Motion or photogrammetry. This increasingly popular data is able to map scenes at large scale and thus serving as a reference, for geolocalization and mobile mapping purposes for example. 

Retrieval in 3D point clouds involves finding the most similar point clouds to a given query within a reference database of 3D point clouds. Research on retrieval within 3D point clouds is still in its early stages. Existing solutions mainly focus on the indexing phase ({\it i.e.} the production of point cloud descriptors) rather than the retrieval phase, which involves efficiently finding similar descriptors in the reference. The current state of the art confirms that the main application of retrieval in point clouds is geolocalization. Unlike approaches relying on visual place recognition, 3D LiDAR point clouds have the benefits of being invariant to lighting and weather conditions \cite{lowry2015visual}, making them particularly relevant for this purpose. 

%\textcolor{red}{PHRASE PAS COMPREHENSIBLE: Point clouds retrieval in the context of geolocalization and place recognition, where the need for geolocating systems or data is essential.} 

\textbf{Applications of retrieval in point clouds.}  Improving geolocation accuracy of a system, based on LiDAR acquisitions, has several key applications. For instance, position correction of LiDAR-equipped smartphones in dense urban environments, where urban canyon reduces GNSS accuracy. Point cloud retrieval may also apply for digital twin updates between a reference point cloud and a new acquisition for land surveying or monitoring structural changes over time. This is also relevant for autonomous navigation, where loop closure detection plays a crucial role in reducing drift in Simultaneous Localization and Mapping (SLAM) systems \cite{zhang2024lidar}. This could extend to media verification, fact checking videos for example, by geolocating 3D point clouds reconstructed from video (e.g with Structure-From-Motion). 

\begin{figure}
    \centering
    \includegraphics[scale=0.55]{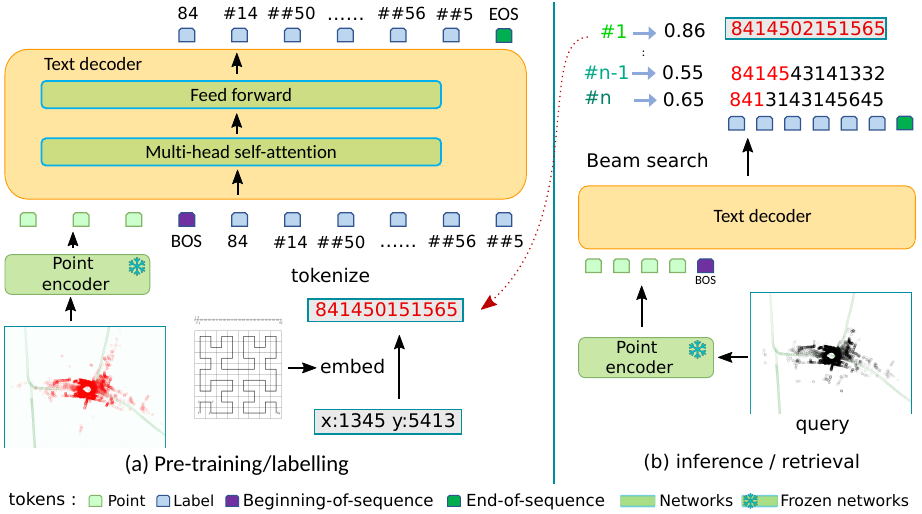}
    % \includesvg[scale=0.8]{dessin.svg}
    \caption{Differentiable search index pipeline \cite{tay_transformer_2022} adapted for 3D data retrieval using GIT architecture \cite{wang_git_2022}. (a) Labeling step: for each input of the database (red point cloud), the text decoder learns to generate the label according to the encoded point cloud. (b) Retrieval: given a point cloud as a query (black point cloud), a beam search is performed by the text decoder to provide the $n$ most probable solutions.}
    \label{fig:enter-label}
\end{figure}

The main objective of the article is to contribute to the retrieval phase, by proposing a scalable retrieval method, facing the size of the reference dataset and the dimensionality of the descriptors. We explore the application of Differentiable Search Index (DSI) \cite{tay_transformer_2022} methods, originally developed for Information Retrieval in large text corpora, to the domain of 3D point clouds. DSI employs a single T5 (text-to-text transfer transformer) model to map string queries to relevant document identifiers (docids). To bridge the gap between the 3D point cloud queries and the textual input required for Information Retrieval, we explored existing Place Recognition methods, such as PointNetVLAD \cite{uy_pointnetvlad_2018} and LoGG3D-Net \cite{vidanapathirana_logg3d-net_2022}. These methods have shown promising results in loop detection by encoding a point cloud into a global descriptor, a single feature vector that summarizes the entire point cloud geometry. Our model takes inspiration from Generative Image-to-text Transformer (GIT)  \cite{wang_git_2022}  which is a Vision Transformer (ViT) used for image captioning. Fig. \ref{fig:enter-label}  illustrates the whole process, where the GIT image encoder is replaced with a point cloud encoder where the text decoder is trained to map 3D data to unique identifiers. The main contributions of this paper are:
\begin{itemize}
    \item We propose DSI-3D, an adaptation of DSI methods to support scalable retrieval in 3D point clouds by training the model to map a point cloud representation to a corresponding \emph{scene docid}. 
    %\item We adapt Vision Transformer (ViT) methods, such as  Generative Image-to-text Transformer (GIT) \cite{wang_git_2022} by replacing their image encoder for a 3D encoder. 
    \item We introduce novel indexing strategies to enhance document representation, incorporating geocoding and Hilbert curve-based methods.
\end{itemize}

This paper is organized as follows: Section \ref{sec:Related} revisits the state of the art on 3D point cloud retrieval and main concepts, applied to the problem of place recognition. Section \ref{sec:method} introduces our proposal while Sections \ref{sec:setup} and \ref{sec:results} present its evaluation, before concluding in Section \ref{sec:conclusion}.

\section{Related work and main concepts}
\label{sec:Related}

In this section, we present a state of the art on 3D point clouds retrieval in Section \ref{sec:SOTA_3Dpoints}, followed by a closer look at the loss function used by LoGG3D-Net in Section \ref{sec:logg3Dnet}. We present the concepts of Differentiable Search Index in \ref{sec:DSI} and how it can be adapted to 3D data with a Vision-language model in Section \ref{sec:GIT}.

\subsection{Retrieval in 3D point clouds}
\label{sec:SOTA_3Dpoints}

Most of the approaches of retrieval in 3D point clouds encountered are applied to the problem of place recognition. This involves determining the geolocation of a points cloud query from the most similar clouds retrieved in a geolocalized reference. First, the reference dataset is usually divided into multiple smaller samples, each of them being geolocalized and indexed with a descriptor. Second, the same type of descriptor is computed for the query and compared to those of the reference dataset. The best matches yield the most likely locations of the query. 

There are multiple families of representation for 3D point clouds. A first distinction can be made between local and global feature-based strategies. 

% local learned
\textbf{Local features-based.} Local features gather local information, like density or local 3D convolution, about a specific region of the cloud, such as particular points or voxels. These place recognition techniques may suffer from accuracy loss due to viewpoint changes or the need for dense point clouds.  Convolutional Neural Networks (CNNs) have demonstrated strong feature learning capabilities for 2D image data \cite{arandjelovic_netvlad_2016}, but their application to LiDAR-based Place Recognition is challenging due to the sparse  and unordered nature of point clouds. PointNet \cite{qi2017pointnet} achieves direct features learning from raw point cloud, while PointNet++ \cite{qi2017pointnet++}, extends this approach by recursively applying \cite{qi2017pointnet}, learning hierarchical features considering distance metric. 

In contrast, there is also a global features-based approach that represents the entirety of the query with one object, often a vector. This global representation tends to extract global information from the aggregation of local features. They have the advantage of being viewpoint-invariant and computationally less expensive for the retrieval.

% global- handcraft
\textbf{Global features handcrafted-based.} Handcrafted-based features do not rely on a previous model learning to be applied for place recognition. Among them, we can cite Scan Context \cite{kim2018scan} which encodes a scene into one rotation-invariant global descriptor using a polar projection of visible points, without requiring prior training. However, this approach struggles with lateral offsets invariance. Scan Context++ \cite{kim2021scan} addresses this issue by employing polar and Cartesian context.

% global- learned
\textbf{Global learned features-based.} By contrast with handcrafted methods, learned point clouds descriptors have benefited from deep learning. A pioneering end-to-end learning-based method, PointNetVLAD \cite{uy_pointnetvlad_2018}, generates a global descriptor from point cloud data. It combines PointNet \cite{qi2017pointnet} for local feature extraction, with NetVLAD \cite{arandjelovic_netvlad_2016}, a learnable CNN layer that aggregates local features into a single global descriptor. Since NetVLAD is a symmetric function, {\it i.e}, it is input order invariant, so can be applied to unordered point clouds. 

% global graph
\textbf{Global features graph-based.} To better capture structural information in large-scale environments, graph-based methods model the relationship between an object and its topology. LPD-Net \cite{liu2019lpd} introduces graph-based aggregation to analyze spatial distribution. It computes ten types of local features, including curvature changes, local point density, and 2D scattering, for each point. A Graph Neural Network (GNN) is then used to aggregate points distributions and local features into a feature space and a Cartesian space respectively. Finally, NetVLAD \cite{arandjelovic_netvlad_2016} generates the global descriptor. While LPD-Net improves retrieval performances upon previous results, it may be limited by its high computational cost and architectural complexity.

% global segment based
\textbf{Global features segment-based.} Segment-based methods divide a scene into meaningful regions (e.g., roads, buildings, vegetation), providing a more structured representation that captures topological relationships between segments, rather than treating the entire point cloud as a single entity. Locus \cite{vidanapathirana_locus_2021} aggregates these segment-based representations into a global descriptor, by modeling the topological relationships and temporal consistency segment feature. This is achieved via second-order pooling, where local features are represented as a matrix, by using the outer product with its transposed matrix and taking the element-wise maximum. It is followed by Eigen-value Power Normalization (ePN) \cite{li2017second} is applied to enhance descriptor description.
%, which decomposes the feature matrix into singular values and raises them to a power $\alpha$. 

%$F^{O_2}$
%\begin{equation}
%    F^{O_2} = \{F^{O_2}_{xy}\}, F^{O_2}_{xy} = \max_{s\in S} f^{O_2}_{xy}(s)
%    \label{sop}
%\end{equation}
%Where $F^{O_2}$ is a matrix with elements $F^{O_2}_{xy}$ $(1 \le x,y \le d)$, $d$ the dimension of the global descriptor and $f^{O_2}(pi) = f(pi)f(pi)^{T} \in \mathbb{R}^{d\times d}$. To enhance descriptor discriminability, Eigen-value Power Normalization (ePN) \cite{li2017second} is applied. It is accomplished with a decomposition into singular values $F^{O_2}$ and raise by a power $\alpha$, where: 

%\begin{equation}
%    F^{O_2}_{\alpha} = U \hat{\lambda} V, \hat{\lambda} = diag(\lambda_{1,1}^{\alpha},\dots, \lambda_{d,d}^{\alpha})
%    \label{epn}
%\end{equation}

% global sparse convo
\textbf{Global features sparse convolution.} 
MinkLoc3D \cite{komorowski_minkloc3d_2021} enhances local features extraction by introducing the first 3D sparse convolution-based architecture. A local feature extraction module is applied to the sparsified point clouds, and a global descriptor is produced using Generalized-Mean pooling \cite{radenovic2018fine}. Similarly, LoGG3D-Net \cite{vidanapathirana_logg3d-net_2022} builds upon sparse volumetric methods, like those in MinkLoc3D, but introduces a local consistency loss function. This function enforces local feature similarity between a query and a positive point cloud (within $3m$), leading to more robust local descriptors. The global descriptor is then obtained using second-order pooling \cite{vidanapathirana_locus_2021}. In Section \ref{sec:logg3Dnet}, we take a closer look at this approach, especially its quadruplet loss, which has inspired our work.

\textbf{Multi-sourced global features.} 
Cross-source 3D place recognition concerns the confrontation of point clouds acquired at different viewpoints, e.g. ground-based LiDAR scans with aerial ones. It presents significant challenges due to differences in coverage, density and noise patterns. To bridge the representation gaps of those types of acquisition, CrossLoc3D \cite{guan_crossloc3d_2023} introduces a cross-source 3D matching approach. It learns a shared embedding space between aerial LiDAR database and ground LiDAR scans. It employs multi-scale sparse convolutions for feature selection, followed by a NetVLAD aggregator to obtain a unique descriptor. An iterative refinement step aligns the two embedding spaces into a shared one. 

%\vspace*{-0.2cm}
\subsection{LoGG3D-Net}
\label{sec:logg3Dnet}
To reinforce the global descriptors similarity during the training, approach LoGG3D-Net applies a quadruplet loss $L_{quad}$ \(\in \left[0, +\infty\right[\), to a tuple ($P_{q}, P_{p}, P_{n}, P_{nbis})$:
\vspace*{-0.2cm}
\begin{multline}
    L_{quad} = 
    \sum_{i}^{N}([\lVert g(P_{q}) - g(P_{p}) \rVert_2^{2}  - \lVert g(P_{q}) - g(P_{n}i) \rVert_2^{2} + \alpha]_{+} \\ 
    + [\lVert g(P_{q}) - g(P_{p}) \rVert_2^{2}
    - \lVert g(P_{nbis}) - g(P_{n}i) \rVert_2^{2} + \beta]_{+}   )
    \label{log_loss}
\end{multline}
where $P_{q}$ is the query point cloud, $P_{p}$ is a positive one within $p=3m$ of the query, $P_{n}$ is a negative point cloud distant from $n=20m$ of the query, $P_{nbis}$ respects the negative condition with both $P_{p}$ and $P_{n}$. g($P_{*}$) is the final global descriptor vector. $N$ is the number of sampled negatives and $[]_{+}$ the hinge loss function, while $\alpha$ and $\beta$ are the loss margin of the quadruplet.
%\textcolor{blue}{Logg3D-Net is trained and evaluated on the KITTI \cite{geiger_vision_2013} and MulRan \cite{kim_mulran_2020} datasets, both of which were collected from moving vehicles in various dynamic urban environments.} 
For the retrieval step, the Cosine distance performs global descriptors retrieval by similarity between the query and the reference dataset items. In \cite{vidanapathirana_logg3d-net_2022}, previous entries adjacent to the query in time by less than $30s$ are excluded from the search to avoid matching with the same instance. 

\subsection{Differentiable Search Index}
\label{sec:DSI}

In text information retrieval, the objective is to find the most relevant document to answer a given question. Paper \cite{tay_transformer_2022} introduces the Differentiable Search Index (DSI), a method that fully parameterized the traditionally multi-stage retrieve-then-rank pipeline within a single T5 transformer model. The DSI architecture also supports beam search,  allowing it to return multiple possible answers ranked by their sequence scores. These scores, or logits \(\in \left]-\infty, +\infty\right[\) represent the raw prediction scores from the language modeling head for each token in the vocabulary before the SoftMax function. The SoftMax function converts them into probabilities over the classes such as SoftMax(logits) \(\in \left[0, 1\right]\). The probability score returned by the T5 model of the generated docid $i$ for a query $q$ can be computed as \cite{zhuang_bridging_2022}: 
\begin{equation}
    p(i|q,\theta) = \prod_{i=1} p(i_{t}| T5_{\theta}(q, i_{0}, i_{1}, ... i_{t-1}))
\label{dsi_prpb}
\end{equation}
where $i_{t}$ is the $i$-th token in the docid string. To achieve this, DSI operates in two fundamental modes:

\textbf{Indexing}: The DSI model learns to associate each document with its corresponding document identifier (docid) using a sequence-to-sequence approach \cite{sutskever2014sequence}. It takes the document tokens as input and generates the corresponding identifier as output. 

\textbf{Retrieval}: For a given input query, the DSI model generates a ranked list of candidate docids through autoregressive generation \cite{de2020autoregressive}. More precisely, three strategies  are used for representing docids in document retrieval:

\textbf{Unstructured Atomic Identifiers} is the simplest approach, where each document is represented by a unique, arbitrary integer identifier. The model is trained to emit one logit which is the last layer unscaled score of the model, for each unique docid, with the SoftMax function applied to the final layer’s hidden state in the decoder. To retrieve the top-$k$ documents, the logits are sorted, and the indices of the highest values are returned. 

\textbf{Naively Structured String Identifiers} represents each document using arbitrary unique integers as tokenized strings, which avoids the need for a large SoftMax layer and simplifies training. A partial beam search tree \cite{sabuncuoglu1999job} is employed to generate the top-$k$ scores. It is similar to a multi-label classification. 

\textbf{Semantically Structured Identifiers} encode semantic information within each document’s identifier, structuring them to reduce the search space with each decoding step. This results in identifiers where semantically similar documents share common prefixes, facilitating retrieval. One method to construct these identifiers is to group similar descriptors by applying $k$-means clustering, then concatenating the resulting cluster indices to form each identifier.

A limitation of \cite{tay_transformer_2022} is the size difference between the indexing phase, which processes an entire text document and the retrieval queries, which are sentence-sized. To address this, \cite{zhuang_bridging_2022} proposes representing each document as a set of relevant queries, using the query generation model docTquery \cite{nogueira2019doc2query}, leading to improved retrieval performances compared to the original DSI model. In our study, the retrieval and the indexing phase are both done on the same format of point clouds descriptors. However query generation could be a way to reinforce the mapping between a scene and its index.   

\subsection{Feature-to-text}\label{sec:GIT}
To bridge the modality gap between text information and 3D points, we first explored existing Vision-Language (VL) models, which are trained on vision-language tasks such as generating image captions. Among image-to-text transformers, GIT stands out for its simple architecture \cite{wang_git_2022}, consisting of a single image encoder and a text decoder. The generated captions are decoded from the sequence of tokens with the highest scores. Decoding starts with a [BOS] (Beginning-Of-Sequence) token and proceeds in an autoregressive manner until reaching the [EOS] (End-Of-Sequence) token or the maximum sequence length. 

%\textcolor{blue}{The image encoder is based on the Florence foundation model, a hierarchical Vision Transformer that includes a linear layer and a layer normalization (LayerNorm) layer. This encoder outputs a 2D feature map, which is then flattened to serve as input to the text decoder.(a supprimer ?)} 
%\textcolor{blue}{The text decoder is a transformer module composed of multiple transformer blocks, each containing a self-attention layer and a feed-forward layer.} 
 
%\textcolor{blue}{The image features are concatenated with the text embeddings as input to the transformer module.}

%\textcolor{blue}{The text decoder is initialized randomly, as \cite{wang2020minivlm} demonstrated that random initialization yields comparable performance to pre-trained weights.}
During pre-training, for each image-text pair, the language modeling $L_{m}$ loss \(\in \left[0, +\infty\right[\) is applied, with $I$ the image,  $y_{i}$ for $i \in \{1,···,N\}$ the text tokens, $y_{0}$ the [BOS] token and $y_{N+1}$ the [EOS] one:
\begin{equation}
    L_{m} = \frac{1}{N+1}\sum_{i}^{N+1} (CE(y_{i}, P(y_{j}| J=0, ..., i-1 ))
\label{git_loss}
\end{equation}
where CE is the cross-entropy loss:
\begin{equation}
   CE = - \sum_{c=1}^{C} \omega_{c} Log\frac{\exp{(x_{c})}}{\sum_{i=1}^{C}\exp{(x_{i})}}.y_{c}
\label{crossentr}
\end{equation}

and $x$ is the input logit, $y$ the target, $\omega$ the weight, and $C$ the number of classes.

\section{Proposed Method}
\label{sec:method}

Inspired by the concepts of Section \ref{sec:Related} and particularly the DSI search method \cite{tay_transformer_2022}, this work aims to bring the scalable aspect of text information retrieval to 3D point clouds retrieval.  

\subsection{DSI-3D}\label{sec:DSI-3D}

 %\textcolor{blue}{To ensure broad applicability to data acquired through various processes (e.g., LiDAR, Structure-from-Motion), the features used for geolocation rely solely on geometric structure, such as the coordinates of points within the scene.} The process begins with a 3D frozen encoder that generates a descriptor of the point cloud. 
Applying the DSI method requires converting 3D data into text form. This involves a two-step process: first, encoding the 3D point cloud into an embedding, and then passing this embedding through a multimodal transformer to generate captions. For the point cloud encoding, we choose LoGG3D-Net \cite{vidanapathirana_logg3d-net_2022} due to its local consistency loss, which significantly enhances retrieval performances for place recognition, but other point cloud global descriptors could be used.
%The encoder is kept frozen to ensure that the DSI model alone learns to associate each point cloud with a specific representation. (\LC{pas exactement on pourrait apprendre les deux en même temps mais trop compliqué! donc la il vaut mieux ne rien dire}
%Generative multimodal models like GIT, learn to associate captions with image descriptors. 
The multimodal transformer chosen is GIT \cite{wang_git_2022} for its streamlined architecture (see Section \ref{sec:GIT}). By substituting the image encoder of GIT with a 3D encoder, we adapt it to generate captions that serve as docids for point clouds, derived from scene descriptors. In order to reinforce the training by increasing the similarity between positive pairs as in LoGG3D-Net, we use both the regular GIT loss and a variation of the language modeling $L_{m}$ loss \eqref{git_loss} to be triplet and quadruplet losses $L_{tripl}$ and $L_{quad}$ (both in $\left[0, +\infty\right[$). These losses are computed using tuples of point clouds: $(P_q, P_p, P_n)$ for the triplet one, and $(P_q, P_p, P_n, P_{nbis})$ for the quadruplet one, as follows:

\begin{equation}
    \begin{aligned}
        L_{tripl} = & \, L_{m}  
        + (L_{mpos} - L_{mneg} + \alpha) \\
    \end{aligned}
\label{tripl_loss}
\end{equation}

\begin{equation}
    \begin{aligned}
        %L_{quad} = & \, L_{m}  
        %+ (L_{mpos} - L_{mneg} + \alpha) \\
        %& + (L_{mpos} - L_{mnegbis} + \beta)
        L_{quad} = &\,  L_{tripl} + (L_{mpos} - L_{mnegbis} + \beta)
    \end{aligned}
\label{quad_loss}
\end{equation}

With $L_{mpos}$, $L_{mneg}$ and $L_{mnegbis}$ representing the cross-entropy loss between the docid of a query point cloud  $P_q$ and output logits for a positive point cloud $P_p$, a negative point cloud $P_n$, and a second negative point cloud  $P_{nbis}$, respectively. $P_{nbis}$ forms a negative pair with both the query $P_q$ and the first negative $P_n$ ; $\alpha$ and $\beta$ are resp. fixed at 0.5 and 0.3. Finally, the retrieval step is performed with a beam search with a restricted vocabulary to meaningful indexes.

\subsection{Docid representation}
For the representation of the docid, we exclude \textit{Unstructured Atomic identifiers} (see Section \ref{sec:DSI}), as they are the least scalable option. For \textit{Naively Structured String identifiers}, a natural analogy can be drawn with the naming of each point cloud in classical benchmarks with acquisition sequences (e.g. KITTI for mobile mapping), where each point cloud is named sequentially from 0 to $N$. The \textit{Semantically Structured identifiers} strategy is also assessed. In this approach, neighboring point clouds with similar descriptors, and thus being in the same cluster, will share similar identifier prefixes. Naturally, this method extends well to revisits.

Additionally, we introduce \textit{Positional Structured identifiers}, a novel representation tailored for place recognition where the coordinates of a point cloud are defined in a Cartesian coordinate system, denoted as %$X_{N}$=\{$x_{0},x_{1},..x_{N-1}$\}, $Y_{N}$=\{$y_{0},y_{1},..y_{N-1}$\} and $Z_{N}$=\{$z_{0},z_{1},..z_{N-1}$\}
$X_{N}$=\{$x_{0},\dots,x_{N-1}$\} and $Y_{N}$=\{$y_{0},\dots,y_{N-1}$\} (the $Z_{N}$ coordinate being excluded here as we focus on planar geolocation), corresponding to the set of strings values of the center coordinates of a sequence of point clouds of length $N$. We propose generating unique identifiers by interlacing the digit strings coordinates of each point cloud $n$, $x_{n}$ and $y_{n}$, starting from the most significant digit of $x_{n}$ down to centimeter precision of $y_{n}$. The values of the coordinates $X_{N}$ and $Y_{N}$ are shifted to include only non-negative values and scaled to ensure centimeter-level accuracy, enabling unique identifiers even for closely revisited locations.

%\begin{algorithmic}
%\\STATE \textbf{Shift coordinates} to ensure non-negative values:
%\\[
%\x_{i}' \gets x_{i} - X_{\text{min}}, \quad y_{i}' \gets x_{i} - Y_{\text{min}}
%\\]

%\\STATE \textbf{Scale coordinates} to centimeter precision:
%\\[
%\x_{i}'' \gets  s \cdot x_{i}', \quad y_{i}'' \gets s \cdot y_{i}' 
%\\]
%\where $s = 100$.

%\\STATE \textbf{Convert coordinates to strings of digits} of $x_{\text{digits}}$ and %\$y_{\text{digits}}$:
%\\[
%\x_{\text{d,i}} \gets \text{String of } x_{i}'', \quad  y_{\text{d,i}} \gets \text{String of } %\y_{i}''
%\\]
%\If necessary, the shorter string are padded with zeros to match the length of the other.

%\\STATE \textbf{Interlace the digits} of \(x_{\text{d,i}} = \{x_{\text{d,i,D}}, %%\x_{\text{d,i,D-1}}, \dots, x_{\text{d,i,0}}\}\) and \(y_{\text{d,i}} =  \{y_{\text{d,i,D}}, y_{\text{d,i,D-1}}, \dots, y_{\text{d,i,0}}\}\) in descending order of significance:
%\[
%\text{Docid}_{i} = x_{\text{d,i,D}} \oplus y_{\text{d,i,D}} \oplus x_{\text{d,i,D-1}} \oplus %y_{\text{d,i,D-1}} \oplus \dots \oplus x_{\text{d,i,0}} \oplus y_{\text{d,i,0}}.
%\]

%where $\oplus$ denotes concatenation, and D is the maximum number of digits across the coordinate strings..
%\end{algorithmic}
As an example, the docid returned for a point cloud $n$ with center coordinates $x_{n}$='1111' and $y_{n}$='2222' will be $docid$='12121212'. This bijective mapping between point clouds and identifiers guarantees that each revisit of a scene will have a closely related identifier.
We also propose a variant of this positional indexing, using space-filling curves, which map a multidimensional grid space onto a 1D curve, such as the Hilbert curve, known for its superior locality preservation \cite{faloutsos1989fractals}. After shifting and rescaling the $X_{N}$ and $Y_{N}$ center coordinates into positive integers, each point cloud center is assigned a unique identifier derived from the Hilbert curve. This identifier corresponds to the distance the curve must travel to reach point $(X_{N}, Y_{N})$, ensuring that the spatially close points have similar index representations. Empirically, for our experiments with the KITTI dataset, we apply the Hilbert curve with 17 iterations, resulting in a grid resolution of $2^{17} \times 2^{17}$.

\section{Experimental Setup}
\label{sec:setup}

In this section, we present the dataset, the evaluation criteria and the experimental setup chosen. Here, LoGG3D-Net serves both as our 3D encoder for DSI-3D and as a baseline for comparison. 

\subsection{Dataset}
% To compare our approach with existing Place Recognition  methods, we evaluate on two LiDAR datasets, KITTI and MulRan, both collected from moving vehicles in dynamic urban environments. We trained and evaluated our model with KITTI on the 6 sequences containing revisits (00, 02, 05, 06, 07, and 08). % and with MulRan on Kaist, DCC03 and Riverside02.  
To ensure comparability with the state of the art, we evaluated our approach for place recognition using the classical KITTI LiDAR datasets \cite{geiger_vision_2013}. We trained and evaluated our model on the six sequences containing revisits (seq. 00, 02, 05, 06, 07, and 08) to assess the robustness of each variant of our method. These sequences represent a total of 18236 unique scenes, divided as follows according to the sequence: 4541, 4661, 2761, 1101, 1101 and 4071 point clouds, covering approximately 1.13km². To verify the scalability of our model, we constructed a new sequence, numbered 22, by concatenating point clouds from the original six ones. Since each sequence is defined in local coordinates, we applied a shift of ($i \times 1000$m, $i \times 1000$m,) to each point cloud coordinates, where $i$ represents the sequence number during the combination process, to prevent overlapping sequences. $80\%$ of each sequence is allocated to the training set, while the remaining 20\% is equally divided between the validation and evaluation sets. Point clouds are assigned to these sets based on their \textit{Naively Structured String identifiers} as follows: those with indices not divisible by five are placed in the training set; those with indices divisible by ten are assigned to the evaluation set; and the remaining point clouds (indices divisible by five but not by ten) are included in the validation set. This division creates a gap every five point clouds, allowing the model to be evaluated on its ability to identify the closest neighbors of scenes it was not trained on. During LoGG3D-Net evaluation, the point clouds in the training set are computed and stored into a 'seen place' database to replicate the parametrization of the training set of our model.

\subsection{Evaluation Criteria and baseline}
Two main criteria are used to evaluate DSI-3D and the baseline:
\begin{itemize}
    \item The $Hits@N$ metric, used in information retrieval, reports the proportion of correct docid ranked within the top $N$ predictions. Here, we focus on the more discriminative case with $N=1$. 
    \item From a place recognition perspective, we evaluate encoding performance based on the maximum $F1$ score or $F1_{max}$ for each sequence \cite{vidanapathirana_locus_2021,vidanapathirana_logg3d-net_2022}. For LoGG3D-Net, it is found by iterating over the similarity threshold that maximizes the $F1$ score. We adapt this $F1$ score threshold for our model, by iterating over logit score, while maintaining the same geometric revisit criteria as in \cite{vidanapathirana_logg3d-net_2022}. A top-1 retrieval is considered positive if its logits scores is below the maximum confidence threshold. A positive retrieval is classified as true positive if it is within $3m$ of the ground truth pose of the query, and as false positive if it exceeds $20m$. For negative retrievals, where the logits exceed the maximum confidence threshold, a true negative is assigned if the query is not revisited, meaning no point cloud within $3m$ of the location of the query is observed with a timestamp greater than $30s$; otherwise, it is considered as a false negative.
\end{itemize}

As baseline, we compare DSI-3D facing descriptor LoGG3D-Net exploited within two retrieval strategies: first classically for exact retrieval based on the exhaustive comparison of all the descriptors in the reference and secondly, more optimally for approximate retrieval based on approximate nearest neighbor (ANN) \cite{vector_DB2024} relying on Locality Sensitive Hashing (LSH) from the FAISS library \cite{douze2024faiss}. More precisely, we used the binary flat LSH index, where the number of bits $H = \in \{32, 256\}$ corresponds to the number of random hyperplane projections (or random rotations in this implementation), respectively, named LSH$_{32}$ and LSH$_{256}$ in the experiments.

\subsection{Implementation Details}
\label{ssec:Implementation }

The experiments are conducted using a PyTorch implementation on a single V100 GPU. For training the DSI model, we use  the pre-trained 3D encoder from LoGG3D-Net and the text decoder from the base-sized GIT checkpoint, fine-tuned on COCO\footnote{https://huggingface.co/microsoft/git-base-coco}. The evaluation is done using the public implementation of LoGG3D-Net and the checkpoints exploited for each sequence were selected using the leave-one-out strategy. For the combined sequence 22, we employed the LoGG3D-Net checkpoint trained while leaving out sequence 00, ensuring that all 18236 scenes were encoded within the same representation space. The best GIT checkpoints are retained for evaluating each sequence and indexing strategy, based on their performance on the validation set. Positive and negative point cloud pairs are sampled based on distances set to $p=3$ and $n=20m$ during the training. Setting $n=20m$ during the evaluation leads to undiscriminating results between the different approaches. A solution to distinguish the best model was to set $n=3m$ during the evaluation. This stricter threshold highlights the ability of the model to identify neighbors close to the query.

% The Logg3D-Net training parameters remain identical to \cite{vidanapathirana_logg3d-net_2022}.
% and from blip2-opt-2.7b\footnote{https://huggingface.co/Salesforce/blip2-opt-2.7b} for Blip2).

\section{Results} \label{sec:results}
%\CL{passer de section à sous section, suppression des redit (trop semblable a la section 4.2}
%To compare our model to a \emph{classic} place retrieval, with direct embedding comparison, we apply our evaluation protocol on both Logg3D-net and our model.
%For Logg3D-Net, instead of sequentially building a database of seen descriptors, comparing it to the current query descriptor, and filtering out descriptors captured less than 30 seconds ago, the database of seen descriptors already contains 80\% of each sequence descriptors. 

% \begin{table}[!htbp]
% \caption{F1 max score of the evaluation on the KITTI dataset with a negative distance threshold of 20 meters}
% \label{table:f1max20m}
% \centering
% \scriptsize
% \begin{tabular}{|c|c|c|c|c|c|c|c|}
% \hline
% \textbf{Model} & \textbf{00} & \textbf{02} & \textbf{05} & \textbf{06} & \textbf{07} & \textbf{08} \\
% \hline
% Logg3D-net &  1.0 & 1.0 & 1.0 & 1.0 & 1.0 & 1.0 \\
% \hline
% Ours (Label) &  &  & 0.9931 & 0.9790 & 0.9926 & 0.9866 \\
% \hline
% Ours (Hierar) &  &  &  & 1.0 & 1.0 & \\
% \hline
% Ours (GPS) & 0.9988 & 1.0 & 1.0 & 1.0 & 1.0 & 1.0 \\
% \hline
% Ours (Hilbert) & 1.0 & 1.0 & 1.0 & 1.0 & 1.0 & 1.0 \\
% \hline
% \end{tabular}
% \end{table}

%We first present the benefits of the quadruplet cross-entropy loss to train a DSI on 3D data. We then evaluate our model and its indexing variants using both an exact and approximate search methods. Finally, we discuss about the advantages of having a retrieval solution at inference time.

This experimental section is organized as follows: Section \ref{sec:loss} evaluates the impact of the loss function on the training of DSI-3D in terms of retrieval performance. Then Section \ref{sec:index} evaluates our models, using different indexing strategies, and compares them to exact and approximate search methods. Finally, Section \ref{sec:complex} discusses the scalability of our approach in terms of query complexity and retrieval time.

\subsection{Loss function}
\label{sec:loss}
Table \ref{table:loss} presents the $F1_{max}$ and $Hits@1$ scores obtained for the six original sequences on our model, with the GPS strategy, by exploiting the triplet and the quadruplet losses of Section \ref{sec:DSI-3D} (again with a negative distance threshold of $3m$). We observe that the triplet loss achieves a marginally higher average $F1_{max}$ score, while the quadruplet one performs slightly better on the $Hits@1$ score. For the remaining experiments, we chose the quadruplet loss, as it provides the best results on both metrics for the longest sequences 00 and 02. % and with shorter training time. 

\begin{table}[!htbp]
\caption{Scores for loss function comparison on DSI-3D}
\centering
\resizebox{\columnwidth}{!}{%
\label{table:loss}
\scriptsize
\begin{tabular}{|c|c|c|c|c|c|c|c|c|c|}
\hline
\textbf{F1max} & \textbf{00} & \textbf{02} & \textbf{05} & \textbf{06} & \textbf{07} & \textbf{08} & \textbf{mean} \\
\hline
$L_{tripl}$ & 0.9775 & 0.9725 & \textbf{0.9946} & \textbf{0.9674} & 0.9954 & \textbf{0.9875} &\textbf{0.9825} \\
\hline
$L_{quad}$  & \textbf{0.9809} & \textbf{0.9758} & 0.9891 & \textbf{0.9674} & \textbf{0.9955} & 0.9851 & 0.9823 \\
\hline
\end{tabular}
}
\vspace{0.3cm}
\centering
\resizebox{\columnwidth}{!}{%
\scriptsize
\begin{tabular}{|c|c|c|c|c|c|c|c|c|}
\hline
\textbf{Hits@1} & \textbf{00} & \textbf{02} & \textbf{05} & \textbf{06} & \textbf{07} & \textbf{08} & \textbf{mean} \\
\hline
$L_{tripl}$ & 0.9560 & 0.9465 & \textbf{0.9892} & \textbf{0.9369} & 0.9820 & \textbf{0.9755} & 0.9644 \\
\hline
$L_{quad}$ &  \textbf{0.9626} & \textbf{0.9507} & 0.9783 & \textbf{0.9369} & \textbf{0.9909} & 0.9705 & \textbf{0.9650} \\
\hline
\end{tabular}
}
\end{table}

%\textcolor{blue}{The quadruplet loss achieves higher scores on both $F1_{max}$ and $Hits@1$ on average and at equivalent training epochs}

\subsection{Indexing strategies comparison}
\label{sec:index}
Table \ref{table:f1max3m} and Table \ref{table:hits1} resp. present the $F1_{max}$ scores and the $Hits@1$ scores obtained for the LoGG3D-Net baseline (without and with index LSH) and several variants of DSI-3D. The names 'Label', 'Hierar', 'GPS' and 'Hilbert' refer to the indexing strategies \textit{Naively Structured String identifiers}, \textit{Semantically Structured identifiers}, \textit{Positional Structured identifiers} and \textit{Hilbert curve Positional Structured identifiers} respectively. Sequence 22 was only trained with the \textit{Hilbert curve Positional Structured identifiers} as it is the most promising indexing strategy.

% \begin{table}[!htbp]
% \caption{$F1_{max}$ scores scores for LoGG3D-Net, LSH and various strategies of DSI-3D.}
% \label{table:f1max3m}
% \centering
% \resizebox{\columnwidth}{!}{%
% \scriptsize
% \begin{tabular}{|c|c|c|c|c|c|c|c|}
% \hline
% \textbf{Model} & \textbf{00} & \textbf{02} & \textbf{05} & \textbf{06} & \textbf{07} & \textbf{08} & \textbf{22}  \\
% \hline
% LoGG3D-Net & 0.9956 & 0.9859 & 1.0 & 0.9954 & 1.0 & 0.9974 &  0.9925 \\
% \hline
% LoGG3D-Net + $LSH_{256}$ & 0.9945 & 0.9792 & 0.9946 & 0.9815 & 0.9626  & 0.9888 &  0.9850 \\
% \hline
% LoGG3D-Net + $LSH_{32}$ & 0.9416 & 0.8979 & 0.9681 & 0.9051 & 0.9670 & 0.9448 & 0.9290 \\
% \hline
% Ours (Label) & 0.8991 & 0.8774 & 0.7906 & 0.7954 & 0.8844 & 0.9044 &  /\\
% \hline
% Ours (Hierar) & 0.9635 & 0.9316 & 0.9684 & 0.9626 & 0.9817 & 0.9684 & /\\
% \hline
% Ours (GPS) &  0.9809 & 0.9758 & 0.9891 & 0.9674 & 0.9955 & 0.9851 & /\\
% \hline
% Ours (Hilbert) & 0.9933 & 0.9781 & 0.9909 & 0.9817 & 0.9955 & 0.9913 & \textcolor{blue}{0.9830} \textcolor{red}{0.9819} \\
% \hline
% \end{tabular}
% }
% \end{table}

\begin{table}
\caption{$F1_{max}$ scores for LoGG3D-Net (with and without LSH) and various strategies of DSI-3D.}
\label{table:f1max3m}
\centering
\resizebox{\columnwidth}{!}{%
\scriptsize
\begin{tabular}{cccccccc}\toprule

\textbf{Model} & \textbf{00} & \textbf{02} & \textbf{05} & \textbf{06} & \textbf{07} & \textbf{08} & \textbf{22}  \\ \midrule
LoGG3D-Net & \textbf{0.9956} & \textbf{0.9859} & \textbf{1.0} & \textbf{0.9954} & \textbf{1.0} & \textbf{0.9974} &  \textbf{0.9925} \\
\midrule
LoGG3D-Net + LSH$_{256}$ &\textbf{0.9945} & \textbf{0.9792} & \textbf{0.9946} & 0.9815 & 0.9626  & 0.9888 &  \textbf{0.9850} \\
LoGG3D-Net + LSH$_{32}$ & 0.9416 & 0.8979 & 0.9681 & 0.9051 & 0.9670 & 0.9448 & 0.9290 \\
Ours (Label) & 0.8991 & 0.8774 & 0.7906 & 0.7954 & 0.8844 & 0.9044 &  /\\

Ours (Hierar) & 0.9635 & 0.9316 & 0.9684 & 0.9626 & 0.9817 & 0.9684 & /\\

Ours (GPS) &  0.9809 & 0.9758 & 0.9891 & 0.9674 & \textbf{0.9955} & 0.9851 & /\\
Ours (Hilbert) & 0.9933 & 0.9781 & 0.9909 & \textbf{0.9817} &\textbf{ 0.9955} & \textbf{0.9913} &  0.9819 \\
\bottomrule
\end{tabular}
}
\end{table}

%vspace{-0.5cm}
\begin{table}[!htbp]
\caption{$Hits@1$ scores for LoGG3D-Net (with and without LSH) and DSI-3D.}
% This score only depends on the positive threshold distance $p$ and does not rely on the LoGG3D-Net distance and on confidence threshold.
\label{table:hits1}
\centering
\resizebox{\columnwidth}{!}{%
\scriptsize
\begin{tabular}{ccccccccc}\toprule
\textbf{Model} & \textbf{00} & \textbf{02} & \textbf{05} & \textbf{06} & \textbf{07} & \textbf{08} & \textbf{22}\\ \midrule
LoGG3D-Net & \textbf{0.9956} & \textbf{0.9722} & \textbf{1.0} & \textbf{0.9910} & \textbf{1.0} & \textbf{0.9950} & \textbf{0.9852}\\
\midrule
LoGG3D-Net + LSH$_{256}$ & \textbf{0.9890} & \textbf{0.9572} & \textbf{0.9892} & 0.9550 & 0.9820 & 0.9755 &  \textbf{0.9693} \\
LoGG3D-Net + LSH$_{32}$  & 0.6110 & 0.5225 & 0.6173 & 0.5676 & 0.8108 & 0.6176 & 0.5049 \\
Ours (Label) & 0.8132 & 0.7645 & 0.8831 & 0.6306 & 0.7928 & 0.8235 & /\\
Ours(Hierar) & 0.9319 & 0.8630 & 0.9387 & 0.9279 & 0.9640 & 0.9387 &/\\
Ours (GPS) & 0.9626 & 0.9507 & 0.9783 & 0.9369 & 0.9909 & 0.9705 &/ \\
Ours (Hilbert) & 0.9846 & 0.9550 & 0.9819 & \textbf{0.9640} & \textbf{0.9910} & \textbf{0.9828} & 0.9644  \\
\bottomrule
\end{tabular}
}
\end{table}

Regardless of the evaluation criteria (Tables \ref{table:f1max3m} and \ref{table:hits1}), we first observe that accelerating LoGG3D-Net retrieval with LSH affects the quality of the responses more or less, compared to exact retrieval (first line in each Table), depending on the value chosen for $H$.

Secondly, among the various indexing strategies tested for DSI-3D, Hilbert curve indexing yields the best results. This leads to the conclusion that the strategies of the state of the art ('Label' and 'Hierar') can be outdated for retrieval dedicated to place recognition. This version does not achieve the same performance as the exact baseline LoGG3D-Net, which provides a slightly better $Hits@1$ score, by an average improvement of 1.6\% (median at 1.3\%) ; for $F1_{max}$, 'Hilbert' is less efficient of 0.8\%  on average (median at 0.5\%). However, we observe the same behavior, as well as scores comparable to DSI-3D ones, for approximate retrieval when adding LSH$_{256}$ to LoGG3D-Net. Globally, DSI-3D with Hilbert curve indexing achieves even a higher average $F1_{max}$ than LoGG3D-Net with LSH$_{256}$ by 0.4\% (median at 0.6\%) and a better average $Hits@1$ by 0.1\% (median at 0.6\%).

%\textcolor{red}{L'article se focalise sur les temps de recherche et pas d'entraînement, je retirerais la phrase suivante, c'est à mon avis tendre le bâton pour se faire battre. The training of sequence 22 took approximately 12 days to achieve a $Hits$ score of 0.96\%. }

%\textcolor{red}{Pas bien convaincue par les arguments de ce paragraphe.... A reréfléchir encore...} \textcolor{blue}{moi non plus.., la "chute" liée aux hits scores est plutôt due au fait que l'entrainement a ralenti trop tôt et qu'il était de plus en plus long et couteux d'augmenter les scores en partant des checkpoints précédents, sans adopter une stratégie de finetuning. Ici, je voulais plutôt prendre le point de vue du F1max que malgré le hits score, on arrive à avoir un bon F1max parce que les cas "faux" sont détectés par le modèle et correctement classifiés.} \textcolor{red}{Through sequence 22, we also observe that DSI-3D seems to scale better according to $F1_{max}$, implying that logits could be interpreted as valuable confidence scores. By further optimizing hyperparameter such as the learning rate and its decay function, it could lead to a faster, and possibly improved, training. }

In conclusion of this section, the behavior of DSI-3D, in terms of retrieval quality, can be considered at least as effective as a conventional state-of-the-art approximate approach. In the following section, we discuss its superiority facing the questions of complexity and retrieval time.

%We observe from Table \ref{table:hits1} that \textcolor{red}{the baseline LoGG3D-Net} outperforms our best model by an average of 2\%.  

\subsection{Complexity and retrieval time}
\label{sec:complex}
In its current implementation \cite{vidanapathirana_logg3d-net_2022}, retrieval with LoGG3D-Net is performed over the entire reference dataset by sequential comparisons (first line of Tables \ref{table:f1max3m} and \ref{table:hits1}). This makes the retrieval time proportional to the size of the dataset ($n$) with a time complexity $O(n)$. The state-of-the-art solutions listed in Section \ref{sec:SOTA_3Dpoints} proceed in the same way. The similarity search in such high-dimensional spaces can be accelerated with indexes such as the LSH \cite{vector_DB2024}, as experimented in the same Tables (lines 2 and 3). The complexity largely differs from the variant used but is generally dominated by $O(n^{\rho})$ with $0 \leq \rho \leq 1$, under the approximate retrieval assumption. In contrast, DSI-3D relies on model inference with constant $O(1)$ complexity, which is a significant advantage facing large datasets of point clouds. It also works under the approximate retrieval paradigm. However, as we have experimented in the previous section, the retrieval quality remains fully comparable to the ones of other solutions under the same paradigm.

In terms of retrieval time, the exact LoGG3D-Net, in its public implementation, ranges from $0.0003$ to $0.0093s$ in average for all the queries. LoGG3D-Net with LSH$_{256}$ is 5.98 times faster, while LoGG3D-Net with LSH$_{32}$ is 7.20 times faster (but with a poor retrieval quality). 
%However, this speed gain comes at the cost of retrieval quality as LoGG3D-Net with $LSH_{32}$  performs 4.6\% worse on average in $F1_{max}$ (median at 4.3\%) compared to LoGG3D-Net with $LSH_{256}$ and shows a drop of 36.7\% in $Hits@1$ (median at 36.5\%).}

For DSI-3D indexed with \textit{Positional Structured identifiers}, currently retrieval time is 
%\textcolor{red}{$0.45s$} 
$0.09s$, by measuring the time it takes for a descriptor to pass through all GIT layers and return the docid. It is worth noting that the current code implementation used for the article is not yet optimized. Nevertheless, to give an insight on the current scalability of DSI-3D, we fitted a linear equation between the size of the reference datasets $N_{desc}$ and retrieval times $T$ with exact LoGG3D-Net ($T = 0.67 N_{desc} - 0.68$). This allows us to estimate that the inversion point occurs at $N_{desc}$ = %\textcolor{red}{756.4e3} 
135.3e3 descriptors in favor of DSI-3D. According to the  settings used, this amount roughly corresponds to a spatial coverage of 
%\textcolor{red}{50km²} 
10km², at which point DSI-3D would become faster in its current not optimized implementation.

Among the factors also affecting the retrieval time of DSI-3D is the number of beams explored during the beam search, which is fixed at $10$, as well as the indexing strategies, affecting decoding steps. Reducing the number of beams could improve retrieval time but may result in less precise responses. Such an evaluation will be part of future experiments.

\section{Conclusion}\label{sec:conclusion}

In this article, we have presented DSI-3D, a new indexing and retrieval approach for 3D point clouds. It drastically accelerates retrieval, compared to current literature which mainly focuses on point cloud description and performs retrieval following a linear time complexity as experimented here, it can be slightly improved with off-the-shelf index such as LSH. The proposal relies on the adaptation of the Differentiable Search Index, but also introduces two new \textit{Positional Structured identifiers}, fitted for the problem of place recognition. This article has laid the foundations of the approach, exhibiting performances similar to those of the state of the art in terms of retrieval quality. In the future, it will be necessary to continue the experiments, in particular by varying the model parameters further, and also by considering larger datasets to confirm its robustness in terms of the quality of the returned responses, its optimality in terms of response time, following a constant complexity, having been proven.

%A way to improve the robustness of our approach is to implement an reranking strategy, benefiting from the results of the beam search to confirm or reconsider the first answer according to the other predictions.

\section{Acknowledgments}
This work is supported by the French Ministry of the Armed Forces - Defence Innovation Agency (AID). It was performed using HPC resources from GENCI-IDRIS (Grant 2024-AD011014030R1).

\bibliographystyle{plain}
\bibliography{These}
\end{document}